\documentclass[conference]{IEEEtran}

\makeatletter

\usepackage{blindtext}
\usepackage{eso-pic}
\IEEEoverridecommandlockouts
\usepackage{cite}
\usepackage{amsmath,amssymb,amsfonts}
\usepackage{algorithmic}
\usepackage{graphicx}
\usepackage{textcomp}
\usepackage{xcolor}
\usepackage{cleveref}
\usepackage{amsmath}
\usepackage{caption}
\DeclareCaptionFont{mysize}{\fontsize{8}{9.6}\selectfont}
\DeclareMathOperator{\arctantwo}{arctan2}
\def\BibTeX{{\rm B\kern-.05em{\sc i\kern-.025em b}\kern-.08em
    T\kern-.1667em\lower.7ex\hbox{E}\kern-.125emX}}
    
\usepackage{eso-pic}
\newcommand\AtPageUpperMyright[1]{\AtPageUpperLeft{%
 \put(\LenToUnit{0.17\paperwidth},\LenToUnit{-2cm}){%
     \parbox{0.9\textwidth}{\raggedleft\fontsize{8}{11}\selectfont #1}}%
 }}%
\newcommand{\conf}[1]{%
\AddToShipoutPictureBG*{%
\AtPageUpperMyright{#1}
}
}    
    
\begin{document}
\title{\vspace*{1cm} Equivariant Map and Agent Geometry for Autonomous Driving Motion Prediction
}

\author{\IEEEauthorblockN{Yuping Wang}
\IEEEauthorblockA{
\textit{University of Michigan}\\
Ann Arbor, United States \\
ypw@umich.edu}
\and
\IEEEauthorblockN{Jier Chen}
\IEEEauthorblockA{
\textit{Shanghai Jiao Tong University}\\
Shanghai, China \\
adam213@sjtu.edu.cn}
}

\maketitle
\conf{\textit{Preprint}}
\begin{abstract}
In autonomous driving, deep learning enabled motion prediction is a popular topic. A critical gap in traditional motion prediction methodologies lies in ensuring equivariance under Euclidean geometric transformations and maintaining invariant interaction relationships. This research introduces a groundbreaking solution by employing EqMotion, a theoretically geometric equivariant and interaction invariant motion prediction model for particles and humans, plus integrating agent-equivariant high-definition (HD) map features for context aware motion prediction in autonomous driving. The use of EqMotion as backbone marks a significant departure from existing methods by rigorously ensuring motion equivariance and interaction invariance. Equivariance here implies that an output motion must be equally transformed under the same Euclidean transformation as an input motion, while interaction invariance preserves the manner in which agents interact despite transformations. These properties make the network robust to arbitrary Euclidean transformations and contribute to more accurate prediction. In addition, we introduce an equivariant method to process the HD map to enrich the spatial understanding of the network while preserving the overall network equivariance property. By applying these technologies, our model is able to achieve high prediction accuracy while maintain a lightweight design and efficient data utilization.

\end{abstract}

%\copyrightnotice{XXX-X-XXXX-XXXX-X/XX/\$XX.00 ©20XX IEEE}


\section{Introduction}
Autonomous driving hinges on the ability to accurately predict vehicle trajectories, a task that ensures safety and efficiency. Recent advancements in machine learning, particularly deep learning, offer promising avenues to address these challenges. However, existing approaches often falter, ignoring the critical need for equivariant motion prediction, where predictions respect the underlying geometric transformations of the environment. Concurrently, invariant interaction reasoning - preserving the interaction dynamics despite transformations - is paramount in multi-agent scenarios. This research employs EqMotion\cite{xu2023eqmotion}, designed with these principles in mind, as its backbone, and innovatively incorporates agent-equivariant maps. These maps, capturing high-definition features without being tied to a specific agent motion, provide a richer spatial understanding. Together, this combination paves the way for a novel approach in autonomous vehicle motion prediction, bridging theory and practical applicability to craft a more robust and insightful predictive model.

EqMotion\cite{xu2023eqmotion}, a groundbreaking motion prediction model, has made strides in various domains but has not explicitly adapted itself to the specific challenges of autonomous vehicle tasks. Its theoretical foundations in equivariance and invariant interaction reasoning are robust, yet the model lacks tailored features to address the complex dynamics and real-world transformations inherent in autonomous driving. A significant limitation is EqMotion's omission of map integration, leaving a gap in contextual understanding of the environment. Without maps, the model's ability to accurately predict intricate vehicular motions may be constrained. These limitations underscore the necessity for a specialized approach, inspiring this research. By building upon EqMotion's backbone and incorporating HD map features and motion equivariant maps, this work aims to bridge these gaps, forging a novel path in autonomous vehicle motion prediction.

Map information, particularly lane centerlines, plays a crucial role in motion prediction for autonomous vehicles, offering vital context on road geometry and traffic constraints. When comparing representations, vectorized representations, as implemented in models like VectorNet\cite{gao2020vectornet}, provide significant advantages over rasterized image representations typically used in Convolutional Neural Networks (CNNs)\cite{choi2021shared}. Vectorized representations retain the geometric structure and can be manipulated through mathematical transformations, facilitating a more compact and expressive encoding of spatial relationships. This approach enhances generalizability and reduces computational complexity. On the other hand, rasterized representations, while popular in traditional CNN-based models, may lead to a loss of geometric fidelity and incur higher computational costs. The principle of equivariance in vectorized representations further bolsters these advantages by ensuring consistent behavior under geometric transformations, enhancing the model's robustness, and maintaining the essential spatial relationships between objects. By preserving this geometric integrity, equivariant vectorized representations offer a more nuanced and efficient approach for motion prediction in complex driving scenarios.

In this research, we have made significant strides in autonomous vehicle motion prediction by introducing several key contributions. First, we propose a novel framework that effectively leverages the equivariance property of map information, particularly in the context of lane centerlines. By embracing this geometric principle, our framework ensures consistent behavior under spatial transformations, offering a more robust and context-aware model. Second, we employ a transformer model on vectorized map representation to extract the vital lane context. This approach has allowed us to capture intricate spatial relationships within the road environment, leading to more nuanced trajectory predictions. Finally, we employ the EqMotion\cite{xu2023eqmotion} to learn the equivalent agent geometric features and reason the invariant interaction between agents. We conduct our experiments on a autonomous driving dataset and the results not only affirm the efficacy of our approach but also represent a significant advancement in the field, offering a more robust and effective model for motion prediction in autonomous driving.

\section{Related Work}

\subsection{Vehicle Trajectory Prediction}
Autonomous vehicle motion prediction has been a focal area of research, with diverse methodologies and frameworks being proposed with a number of them utilizing deep neural networks\cite{li2021spatio,li2020evolvegraph,li2019conditional}. LSTM-based models like Social-LSTM\cite{alahi2016social} have been utilized for capturing temporal dependencies. Convolutional Neural Networks (CNN) have been employed to process rasterized map data\cite{choi2021shared}. More recently, attention mechanisms like Transformer models have been adapted for motion prediction\cite{ngiam2021scene}. Graph-based approaches like Graph Neural Networks have been explored for multi-agent interaction\cite{cao2021spectral, girase2021loki, sun2022interaction}. VectorNet\cite{gao2020vectornet} has introduced the use of vectorized representations, highlighting the importance of preserving geometric relationships. These diverse approaches lay a rich foundation for ongoing innovations in the field.

\subsection{Map Information Encoding}
Map information encoding is a critical aspect of motion prediction, with two main approaches: rasterized and vectorized representations. Rasterized encoding, which converts map information into pixel grids, has been employed in many CNN-based model\cite{choi2021shared, ma2021multi}. It offers simplicity and compatibility with standard image processing techniques but suffers from loss of geometric precision and can be computationally expensive. In contrast, vectorized representations, like those used in VectorNet\cite{gao2020vectornet}, preserve the geometric structure of the map, enabling accurate encoding of spatial relationships. Vectorized encoding ensures scalability and fidelity but may require more complex processing to fully exploit its potential. Recent advancements in transformer-based models have demonstrated the efficacy of vectorized encoding in autonomous driving applications\cite{jiang2023vad}. The choice between these approaches hinges on specific requirements, balancing simplicity with geometric accuracy.

\subsection{Equivariant Feature Extraction}
The concept of equivariance has become particularly prominent in the realm of 2D image analysis. Owing to the susceptibility of CNN structures to rotational alterations, there has been a pursuit of designs that embrace rotation-equivariance, such as the implementation of directional convolutional filters\cite{marcos2017rotation}. In Graph Neural Networks, \cite{thomas2018tensor} achieves both rotation and translation equivariance by using tensor field neural networks. In these works, equivariant models have shown their advantage of an efficient training process by avoiding data augmentation. A myriad of research initiatives have proposed equivariant layers tailor-made for specific tasks, including protein structure decoding\cite{shi2022protein}. However, many remain limited to state prediction and overlook sequence data. Few attempts, like one that focuses on coordinate processing, strive for motion equivariance. Our model innovates an equivariant map processor on top of the equivariant sequence predictor, EqMotion\cite{xu2023eqmotion}, to predict the trajectories for autonomous vehicles.

\section{Problem Formulation}

In this section, we define the formulation of the motion prediction task, which is to predict the ego agent motion given the past motion of itself and neighbor agents, plus the lane centerlines near the ego agent.

Our first input consists of the historical trajectories of \(N\) agents, including both the ego agent and neighbors. For each agent \(a\) at each timestep \(t\) in the input sequence, we have its location vector \(x_a^t\) being its \(x\), \(y\) coordinates. Thus, we have \(N \times T_{\text{in}}\) coordinate pairs, which we denote as $X \in \mathbb{R}^{N \times T_{in} \times 2}$. Our second input is a \(Q\) lane center lines each represented as \(L\) coordinate points, each also represented by its coordinates \(x\), \(y\). Thus, we have \(Q \times L\) coordinate pairs, which we denote as the \textit{M} (map) matrix and $M \in \mathbb{R}^{Q \times L \times 2}$. Our output $\hat{y}$ is the coordinates for the ego agent over the next \(T_{\text{out}}\) timesteps. Thus, it has \(T_{\text{out}}\) coordinate pairs, $\hat{y} \in \mathbb{R}^{T_{out} \times 2}$.

During training and validation, we have access to the ground truth future trajectory for the ego agent as $y \in \mathbb{R}^{T_{out} \times 2}$. Our learning goal is to have $\hat{y}$ minic $y$ as much as possible.
Note that \(N\), \(T_{\text{in}}\), \(T_{\text{out}}\), \(Q\), \(L\) are fixed configuration parameters. When in the scene we have less than \(N\) total agents or less than \(Q\) centerlines, we'll pad the invalid values in the matrices with 0.

\section{Methodology}
\begin{figure}
    \centering
    \includegraphics[width=60mm,keepaspectratio]{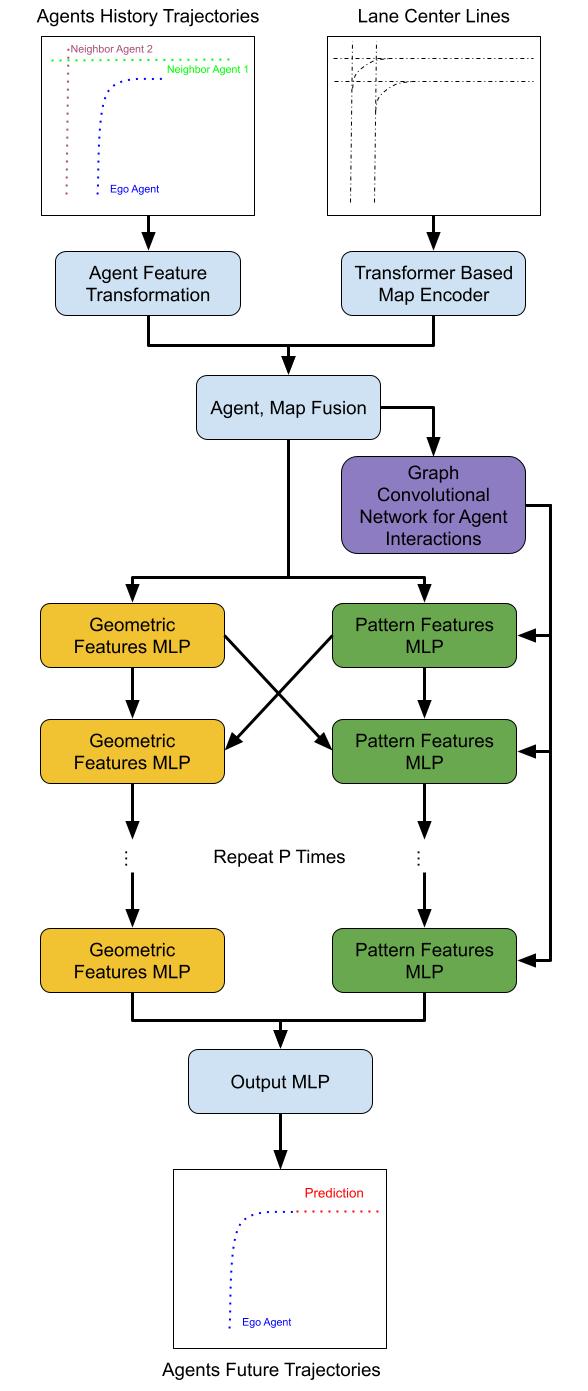}
    \caption{Model Architecture}
    \label{fig:enter-label}
\end{figure}

\subsection{Equivariant Map Feature}

Map features provide the context for our ego motion prediction. To account for equivariant nature between the map and ego agent, we first translate the map coordinate frame to align with the current location at time $t$ of the ego agent $a$, which is denoted as $x_a^t$. For all the center lines and center line points in the map $M_l^q$, $l \in [0, L]$ and $q \in [0, Q]$, we have:
\begin{equation}
M_{\text{centered}, l}^q = M_l^q - x_a^t\label{eq:translatemap}.
\end{equation}
In addition to translating the map, we also rotate the map to align with the ego agent heading which is computed from the current velocity:
\begin{equation}
v_a^t = x_a^{t} - x_a^{t-1},
\end{equation}
\begin{equation}
\theta_a^t=\arctantwo(v_a^t[1], (v_a^t[0])\label{eq},
\end{equation}
\begin{equation}
R_{\theta} = \begin{bmatrix}
\cos(\theta_a^t) & -\sin(\theta_a^t) \\
\sin(\theta_a^t) & \cos(\theta_a^t)
\end{bmatrix},
\end{equation}
\begin{equation}
M_{\text{rotated}, l}^q = R_{\theta} \cdot M_{\text{centered}, l}^q.
\label{eq:rotatemap}
\end{equation}

\subsection{Explanation on Map Equivariance}

We want to show our map transformation is equivariant to agent motion. This means for arbitrary agent movement, the map's coordinates in the agent frame will shift the same amount. We denote each coordinate pair in the map $M$ as $m_i$ and $i \in [0, L \times Q]$. The agent's current position is represented by $x_a^t$ and its heading represented by $\theta$.

\subsubsection{Translation Equivariance}

If the agent moves by a translation vector $b$, its new position is:
\begin{equation}
\tilde{x}_a^{t} = x_a^t + b
\end{equation}

We then have the transformed coordinates of $m_i$ after this translation, relative to the new agent's position, are:
\begin{equation}
\tilde{m_i}  = m_i - \tilde{x}_a^{t} = m_i - (x_a^t + b) = m_i - x_a^t - b.
\end{equation}

This is equivalent to translating the original transformed coordinates $m_i - x_a^t$ by $-b$.

\subsubsection{Rotation Equivariance}

If the agent rotates by an angle $\phi$ in the counter-clockwise direction, the agent's new heading is $-\phi + \theta$. We then have the transformed coordinates of $m_i'$ after this rotation, relative to the agent, are:
\begin{equation}
\check{m_i} = R_{(-\phi + \theta)} * \tilde{m_i} = R_{-\phi}R_{\theta}* \tilde{m_i}.
\end{equation}

This is equivalent to rotating the original rotated map in \ref{eq:rotatemap} in the clockwise direction by $\phi$. 

Thus, the operation of transforming map coordinates to the frame of a moving agent is equivariant with respect to the agent's movement.

\subsection{Map Feature Encoding}
Given the map feature is now equivariant to the agent motion, we now further expand its dimension space for downstream backbone. Specifically, we reshape the rotated map $M_{\text{rotated}}$ so that the lanes and waypoints per lane are vectorized into a vector $M_{\text{vectorized}}$ where $M_{\text{vectorized}, n}$ is a single coordinate pair and $n \in [0, L \times Q]$, which we then encode $M_{\text{vectored}}$ with a Transformer \cite{vaswani2017attention} layer that is composed of the following:

\begin{enumerate}
    \item Multi-Head Self-Attention
    \begin{equation}
        Attention(Q,K,V) = softmax \left(\frac{QK^T}{\sqrt{d_k}} \right)V,
    \end{equation}
    where
    \begin{align}
        Q & = M_{\text{vectorized}} \cdot W_Q \\
        K & = M_{\text{vectorized}} \cdot W_K \\
        V & = M_{\text{vectorized}} \cdot W_V,
    \end{align}
    and $W_Q$, $W_K$, $W_V$ are the matrices for linear transformation. $d_k$ is the dimensionality of the keys in the multi-head attention mechanism.
    
    \item Add and Norm
    \begin{equation}
        Z = LayerNorm(X + Attention(Q, K, V))
    \end{equation}

    \item Feed-Forward Neural Network (FFN)
    \begin{equation}
        FFN(Z) = ReLU(ZW_1 + b_1)W_2 + b_2
    \end{equation}
    \begin{equation}
        Y = LayerNorm(Z + FFN(Z))
    \end{equation}
\end{enumerate}

The resulting vector $Y$ is then concatenated with the agent features as fed to the downstream.
\begin{equation}
    M_{\text{features}} = [X; Y]
\end{equation}

\subsection{Geometric Feature and Pattern Feature Learning}

We here employ the Eqmotion Network \cite{xu2023eqmotion}. Specifically, given our $AgentMapFeatures$ from the above, we obtain the initial geometric $G^0 \in \mathbb{R}^{N \times T_{out}}$ and pattern $H^0 \in \mathbb{R}^{N \times hidden\_dim}$ features:
\begin{equation}
G^0, H^0 = \mathfrak{F}_{\text{FeatureInitLayer}}(M_{\text{features}}).
\end{equation}
We then apply a graph convolution network to infer the relationships between the agents, and the map context:
\begin{equation}
{e_{ij}} = \mathfrak{F}_{\text{GraphConvolutionLayer}}(G^0, H^0).
\end{equation}
With the above, we will sequentially apply two networks to learn the geometric and pattern features in an interleaving fashion. This step will repeat $P$ times and $P$ is a hyperparameter:
\begin{equation}
G^{p} = \mathfrak{F}_{\text{GeometricLayer}}(G^{p-1}, H^{p-1}, {e_{ij}}),
\end{equation}
\begin{equation}
H^{p} = \mathfrak{F}_{\text{PatternLayer}}(G^{p-1}, H^{p-1}).
\end{equation}
For the detailed implementation of the above layers please refer to EqMotion\cite{xu2023eqmotion}.

\subsection{Output Decoder}

We take the final geometric feature $G^P \in \mathbb{R}^{N \times hidden\_dim}$ and decode it with a 4-layer MLP which produces the output trajectory $\hat{y} \in \mathbb{R}^{T_{out} \time 2}$. We will also add the current position of the agent to it:
\begin{equation}
\hat{y} = MLP(G^P - \Bar{G^P}) + \Bar{G^P} + \Bar{X_0}.
\end{equation}

\subsection{Loss Function}

We use the Average Displacement Error as the loss function:
\begin{equation}
\text{ADE} = \frac{1}{T} \sum_{t=1}^{T} \| \hat{y}_t - y_t \|_2.
\end{equation}

\section{Experiments}

\subsection{Dataset and Evaluation Metrics}
In our work, we used the Argoverse motion forecasting dataset\cite{chang2019argoverse}, a comprehensive and well-regarded collection specifically designed for autonomous vehicle applications. Argoverse 1 provides a training set of around 200k scenes, each includes high-definition maps and both ego and neighbor agent trajectories. By leveraging this dataset, we were able to validate and benchmark our model's performance in realistic scenarios, ensuring that the findings are representative and applicable to real-world autonomous driving contexts. In this dataset, agent history is composed of coordinates from the past 2 seconds, and agent futures are the next 3 seconds, both sampled at 10Hz. 

In our comprehensive evaluation, we utilized both Average Displacement Error (ADE) and Final Displacement Error (FDE) metrics at multiple time intervals (1s, 2s, and 3s) to thoroughly assess the predictive capabilities of our model. ADE measures the average difference between predicted and actual trajectories over specified time horizons, providing insights into prediction accuracy. On the other hand, FDE quantifies the endpoint difference between predicted and actual trajectories, offering a distinct perspective on model performance. By employing both ADE and FDE at different time intervals, we gained a comprehensive understanding of our model's accuracy in both intermediate and long-term predictions. This multi-faceted evaluation approach enabled us to identify not only immediate accuracy but also the ability of our model to sustain accurate predictions over longer horizons. These metrics reinforce the rigor of our evaluation process and highlight our commitment to providing reliable and insightful autonomous vehicle motion predictions.

\subsection{Implementation Details}

\begin{figure}
    \centering
    \includegraphics[width=\linewidth]{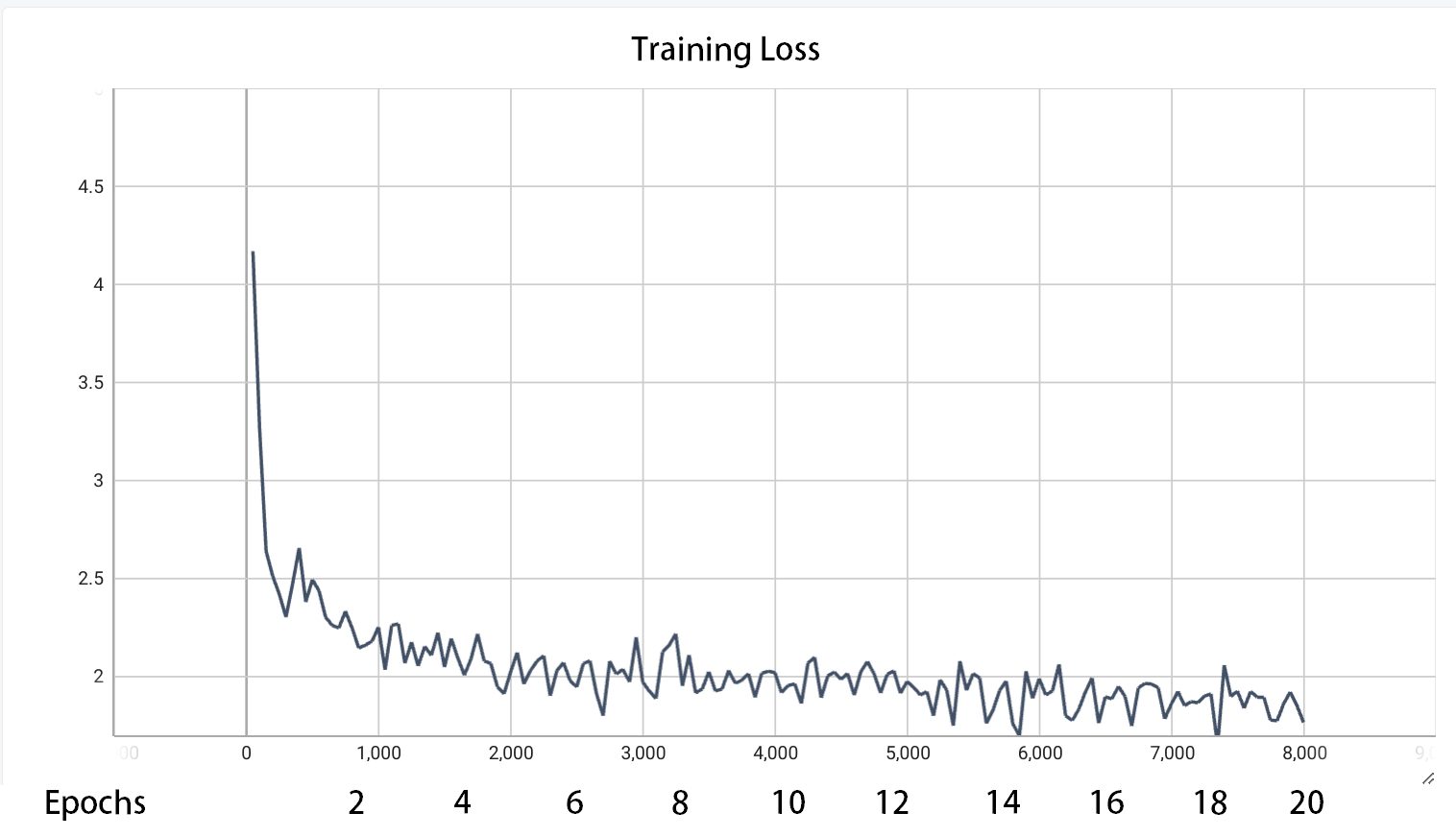}
    \caption{Training Loss Decay}
    \label{fig:train-loss}
\end{figure}

During training, we use the Adam optimizer with a learning rate of $1e^{-5}$. We train our network for 20 epochs and used a batch size of 512. Our training hardware is the NVidia RTX 3060Ti with 8GB of graphic memory. The training loss decay can be seen in \Cref{fig:train-loss}.

In our model, we used the configuration parameters as in \Cref{table:config}.

\begin{table}
\normalsize
\caption{Configuration Parameters}
\label{table:config}
\begin{tabular}{||p{4cm} | p{2cm} | p{1cm}||} 
 \hline
 Description & Name & Value \\ [0.5ex] 
 \hline\hline
Length of input sequence & $T_{in}$ & 20 \\ 
 \hline
Length of output sequence & $T_{out}$ & 30 \\ 
 \hline
Number of total agents in the scene & $N$ & 4 \\
 \hline
Number of lane centerlines in the map & $Q$ & 10 \\
 \hline
Number of points to represent each centerline & $L$ & 100 \\ \hline
Number of repeats on the feature learning layers & $P$ & 20 \\ \hline
Size of all hidden dimension & $hidden\_dim$ & 64 \\ \hline
Number of layers in MLP &  & 4 \\ \hline
Number of heads in transformer & & 12  \\
 [1ex] 
 \hline
\end{tabular}
\end{table}

\subsection{Baseline Model}

To measure the performance of our model, we developed a Long Short Term Memory (LSTM) based network to serve as the baseline. LSTM is a popular network that is able to learn a temporal sequence such as in \cite{alahi2016social} and \cite{jeong2020surround}, in which LSTM is used to predict vehicle behavior at intersections.
\begin{figure}
    \centering    \includegraphics[width=85mm,keepaspectratio]{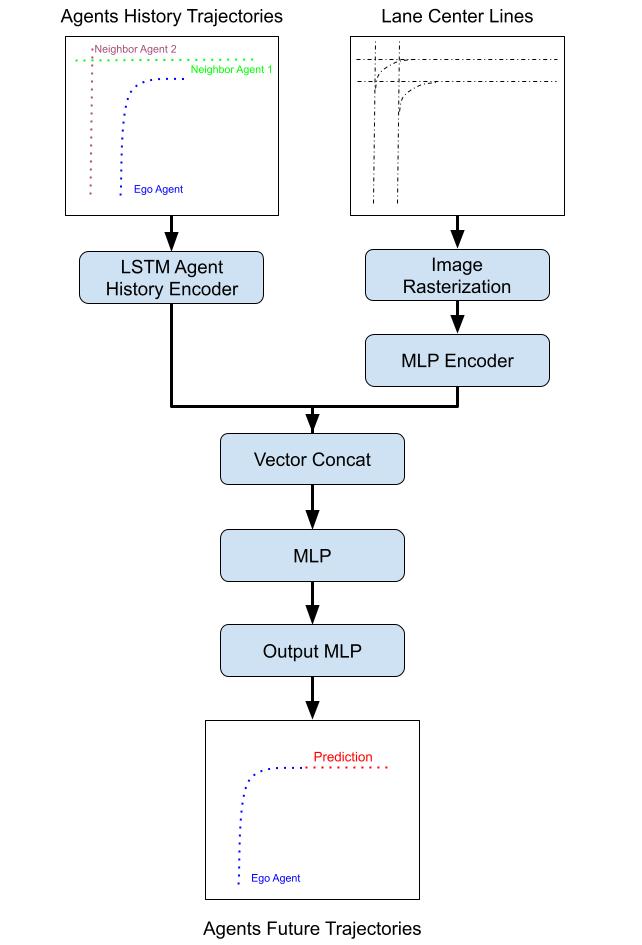}
    \caption{LSTM Baseline Architecture}
    \label{fig:LSTM Baseline Architecture}
\end{figure}
In our LSTM network, we rasterize the map into a $200\times200$ boolean image with the lane centerlines represented as having a value of $1$ in the image while the others have $0$. The image then goes through two convolution neural networks (CNN), each has $3\times3$ kernels followed by a MaxPool layer with stride $2$. The convoluted feature then goes through a 4-layer MLP to produce a feature vector, which we then concatenate to the feature vector produced by the LSTM. Lastly, we use a 4-layer MLP to produce the output trajectory.

\subsection{Comparative Studies}
To provide a better insight into the effectiveness of using the map feature processing and encoding logic via Transformer blocks, we performed experiments with these modifications.
\begin{enumerate}
    \item Skipping the map rotation logic in \cref{eq:rotatemap} and only perform translation.
    \item Replacing the transformer block with a self attention layer\cite{li2021rain} with a graph size of 64.
    \item Replacing the transformer block with the same CNN model as in the baseline.
    \item Ablating the map processing logic to measure the impact of the map information.
\end{enumerate}

% Arranged here for formatting.
% \begin{table*}[t!]
% \centering
% \caption{Comparison of Different Models}
% \label{table:comparison}
% \begin{tabular}{lcccccccc}
% \hline
% Method & ADE at 1s & ADE at 2s & ADE at 3s & FDE at 1s & FDE at 2s & FDE at 3s & Parameters & Training\\
% \hline
% LSTM Baseline & 0.922 & 1.827 & 2.957 & 1.806 & 3.994 & 6.503 & 10.1M & 56min \\
% Our model, no map & 0.602 & 1.037 & 1.667 & 0.982 & 2.182 & 3.715 & 5.8M & 213min\\
% Our model, translated map only, attention & 0.606 & 1.040 & 1.663 & 0.997 & 2.181 & 3.682 & 8.6M & 227min \\
% Our model, rotated map, attention & 0.588 & 1.036 & 1.661 & 0.987 & 2.180 & 3.662 & 8.6M & 224min\\
% Our model, translated map only, transformer & 0.591 & 1.042 & 1.655 & 0.990 & 2.161 & 3.679 & 9.5M & 235min\\
% \textbf{Our model, rotated map, transformer} & \textbf{0.576} & \textbf{1.009} & \textbf{1.617} & \textbf{0.947} & \textbf{2.126} & \textbf{3.604} & \textbf{9.5M} & \textbf{231min} \\
% Our model, rotated map, CNN & 0.812 & 1.320 & 2.051 & 1.252 & 2.64 & 4.41 & 47M  &18hrs\\
% \hline
% \end{tabular}
% \end{table*}
\begin{table*}[t!]
\caption{Comparison of Different Models}
\label{table:comparison}
\centering
\begin{tabular}{lcccccccc}
\hline
Method & ADE at 1s & ADE at 2s & ADE at 3s & FDE at 1s & FDE at 2s & FDE at 3s & Parameters & Training\\
\hline
LSTM Baseline & 0.922 & 1.827 & 2.957 & 1.806 & 3.994 & 6.503 & 10.1M & 56min \\
Our model, no map & 0.602 & 1.037 & 1.667 & 0.982 & 2.182 & 3.715 & 5.8M & 213min\\
Our model, translated map only, attention & 0.606 & 1.040 & 1.663 & 0.997 & 2.181 & 3.682 & 8.9M & 227min \\
Our model, rotated map, attention & 0.588 & 1.036 & 1.661 & 0.987 & 2.180 & 3.662 & 8.9M & 224min\\
Our model, translated map only, transformer & 0.556 & 0.989 & 1.619 & 0.894 & 2.023 & 3.490 & 10.1M & 195min\\
\textbf{Our model, rotated map, transformer} & \textbf{0.549} & \textbf{0.967} & \textbf{1.591} & \textbf{0.895} & \textbf{1.996} & \textbf{3.437} & \textbf{10.1M} & \textbf{211min} \\
Our model, rotated map, CNN & 0.812 & 1.320 & 2.051 & 1.252 & 2.64 & 4.41 & 47M  &18hrs\\
\hline
\end{tabular}
\end{table*}
% \begin{figure}[t]
% \centering
% \begin{tabular}{cc}
% (a) Overlay on Map & (b) Zoom in \\
% \includegraphics[width=0.45\linewidth]{1.png} & \includegraphics[width=0.45\linewidth]{2.png} \\
% \includegraphics[width=0.45\linewidth]{3.png} & \includegraphics[width=0.45\linewidth]{4.png} \\
% \includegraphics[width=0.45\linewidth]{5.png} & \includegraphics[width=0.45\linewidth]{6.png} \\
% \includegraphics[width=0.45\linewidth]{7.png} & \includegraphics[width=0.45\linewidth]{8.png} \\
% \end{tabular}
% \caption{Visualizations}
% \label{fig:Visualizations}
% \end{figure}

\begin{figure*}[!htb]
\centering
\begin{tabular}{cccc}
\includegraphics[width=0.23\linewidth]{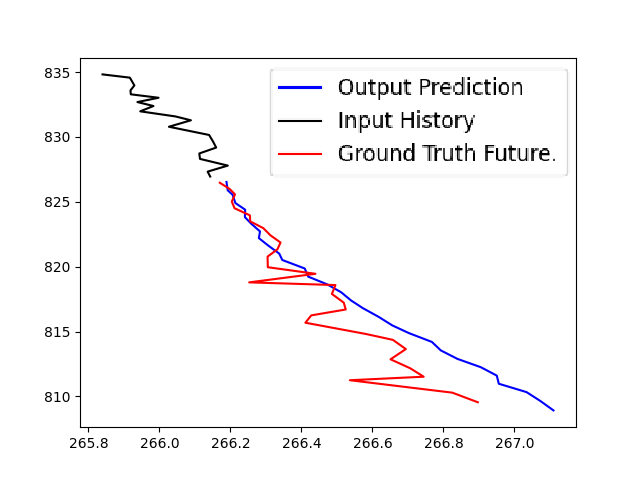} & \includegraphics[width=0.23\linewidth]{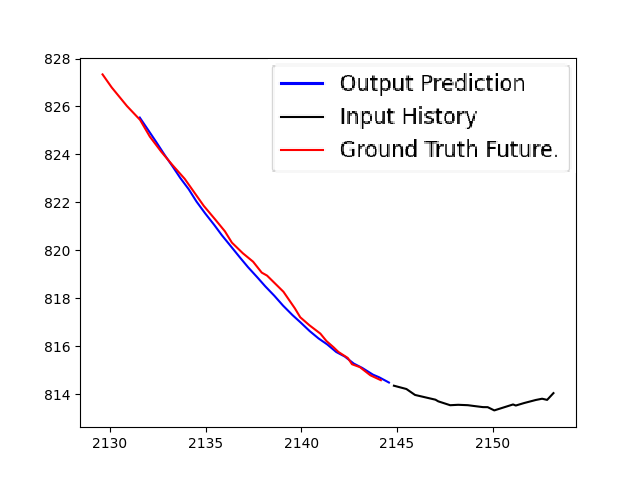} &
\includegraphics[width=0.23\linewidth]{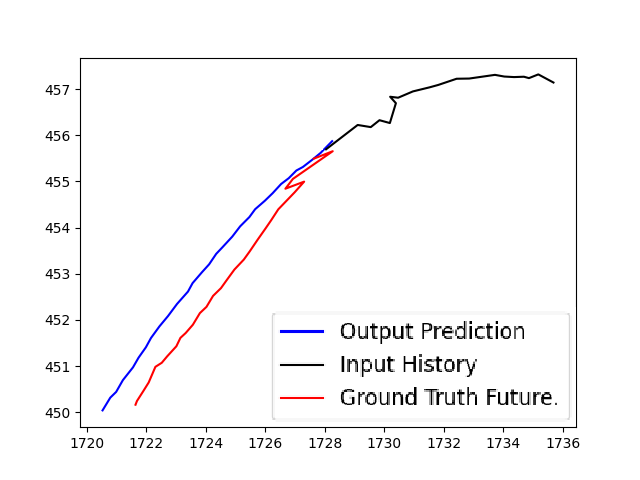} & \includegraphics[width=0.23\linewidth]{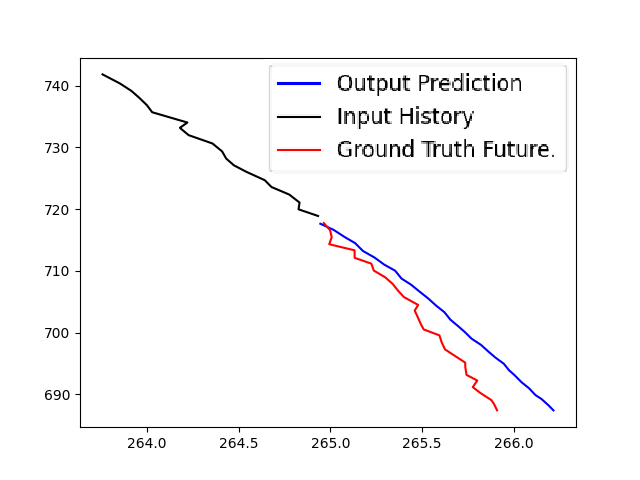} \\
\includegraphics[width=0.23\linewidth]{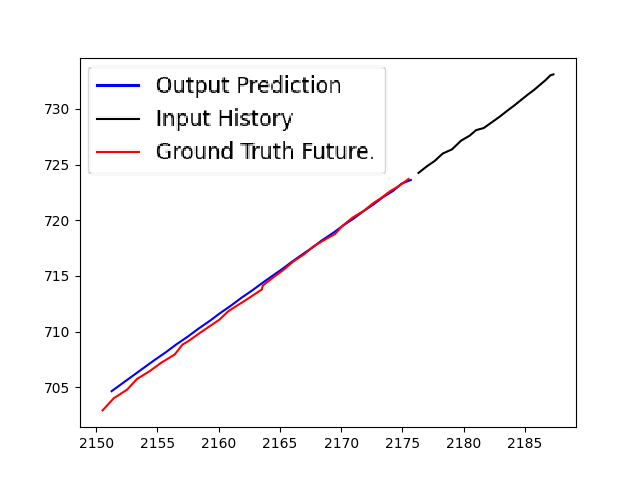} & \includegraphics[width=0.23\linewidth]{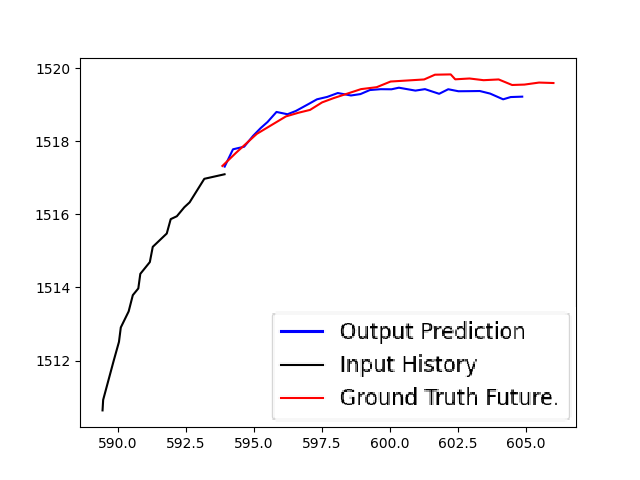} &
\includegraphics[width=0.23\linewidth]{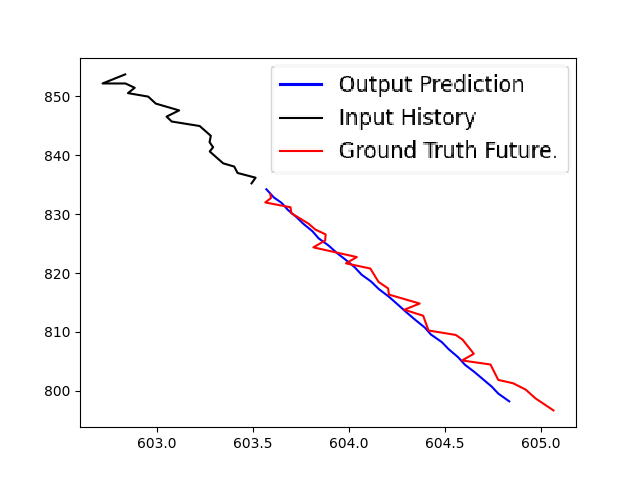} & \includegraphics[width=0.23\linewidth]{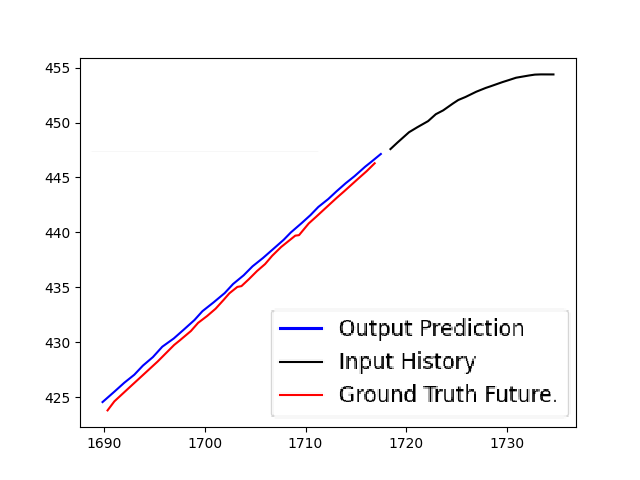} 
\end{tabular}
\caption{Visualizations of the predicted trajectories, agent history input, and ground truth trajectories.}
\label{fig:Visualizations}
\end{figure*}

% Arranged here for formatting.

\subsection{Quantitative Results}
The quantitative results are shown in  \Cref{table:comparison}. Rows 1 - 2 are the results for not using a map. Rows 3 - 7 are the results for using a map.

Our LSTM baseline has higher errors in ADE and FDE, signifying less accurate predictions. It has a moderate model size and reasonable training time.
By replacing the LSTM backbone with EqMotion\cite{xu2023eqmotion}, we see an immediate improvement in both ADE and FDE compared to the baseline and reduced complexity in terms of the number of parameters. Lengthier training time may indicate a more intricate learning process.
We observe that regardless of the map feature encoder used, we can see a performance improvement if we align the map with agent heading. Note that rotating the map does not incur an increase in the number of parameters. Thus it does not affect model size or training time.
Using a transformer model than a simpler self attention one costs 1.2M more parameters, but also better performance.
Lastly, applying a rasterized view of the map and supply it to the model performs worse than the our equivariant map processor, even with many more parameters and longer training time.

In conclusion, our models with map information generally outperform the non map models in prediction accuracy. Different architectural choices like map rotation and attention/transformer showed a noticeable improvement. The model with rotated map and transformer architecture appears to be a overall better choice given models at row 3-6 have similar sizes. Careful consideration is needed for the CNN variant, as it has higher complexity and significantly longer training time. Further experiments with hyperparameter tuning, different feature engineering, or additional data might lead to more robust insights.
\subsection{Qualitative Results}

Some visualizations of our model predictions are shown in \Cref{fig:Visualizations}. We observe the predicted trajectories appear to stay close to the ground truth, suggesting our model is able to predict reasonably well. Additionally, the ground truth data suffers from sampling noise, represented as the spikes in the plot. Our prediction, however, produces a much smoother and realistic trajectory.

\section{Conclusion}
In conclusion, this research presents an innovative integration of EqMotion and HD map features, a first in autonomous vehicle motion prediction. The theoretical rigor and empirical success of this approach promise a transformative impact on the field, leading to more robust, accurate, and generalized motion prediction. The synergy between geometric understanding, interaction reasoning, and environmental mapping sets a new benchmark for autonomous driving technology, with far-reaching implications for the future of transportation.

\bibliographystyle{IEEEtran}
\bibliography{references}

\section{Broader Impacts}

This research introduces a theoretically grounded and practically effective framework for motion prediction that ensures equivariance to Euclidean transformations and invariant agent interactions. Such properties are crucial for building robust autonomous driving systems that generalize well across varied spatial configurations and driving contexts~\cite{wang2025cmp, zhang2021point}. By integrating equivariant HD map features and leveraging EqDrive~\cite{wang2023eqdrive}, our model provides improved prediction accuracy while maintaining a compact and efficient architecture.

Beyond this contribution, our broader research portfolio advances the safety and reliability of autonomous and robotic systems across several fronts. In occupancy prediction~\cite{wang2025uniocc}, multi-agent trust protocols~\cite{li2025safeflow}, aerial-ground collaboration~\cite{gao2025airv2x}, and generative simulation frameworks~\cite{wang2025generative, liu2024toward}, we systematically target core challenges in scalability, contextual reasoning, and inter-agent safety.

We also explore how large vision-language models understand spatial tasks~\cite{xing2025can,xing2025re,xing2025openemma}, releasing benchmarks like MapBench to promote community-driven improvements in embodied intelligence. Additionally, our work on novel RGBD-based view synthesis~\cite{hetang2023novel} contributes to enhanced indoor robotic perception. To encourage reproducibility and open development, we have released supporting datasets and toolkits to the community.

Together, these efforts contribute to the development of safer, more interpretable, and context-aware autonomous agents capable of operating reliably in dynamic real-world environments.

\end{document}